\pdfoutput=1

\documentclass[letterpaper, 10 pt, journal, twoside]{IEEEtran}

\IEEEoverridecommandlockouts

\usepackage{amsmath}
\usepackage{amssymb}
\usepackage[ruled,linesnumbered]{algorithm2e}
\usepackage{graphicx}
\usepackage{booktabs}
\usepackage{url}
\usepackage{xcolor}
\usepackage{microtype}
\usepackage{eso-pic}
\AddToShipoutPictureBG*{%
  \AtPageLowerLeft{\put(0,14){\makebox[\paperwidth]{\parbox[b]{0.9\paperwidth}{\scriptsize\centering
  \copyright~2026 IEEE. Personal use of this material is permitted. Permission from IEEE must be obtained for all other uses, in any current or future media, including reprinting/republishing this material for advertising or promotional purposes, creating new collective works, for resale or redistribution to servers or lists, or reuse of any copyrighted component of this work in other works.}}}}}
\newcommand{\revised}[1]{#1}
\ifdefined\RALDIFF
  \newcommand{\changed}[1]{\textcolor{blue}{#1}}
\else
  \newcommand{\changed}[1]{#1}
\fi

\title{VQActFlow: Vector-Quantized Action Mode Steering for Multi-Task Robot Manipulation}
\author{\changed{Zhigen Zhao, Mark Leggiero, Yipu Chen, Haoran Liu, Yifan Wu, Huishu Xue, Sirui Zhan, and Ye Zhao}%
\thanks{\changed{Manuscript received: June 26, 2026; Revised September 30, 2026; Accepted October 13, 2026.}}%
\thanks{\changed{This paper was recommended for publication by Editor J\'{u}lia Borr\`{a}s Sol upon evaluation of the Associate Editor and Reviewers' comments.
This work was partially supported by General Intelligent Machines, Inc. and the United States Department of Agriculture under Grant 2023-67021-41397.}}%
\thanks{\changed{The authors are with the Institute for Robotics and Intelligent Machines (IRIM), Georgia Institute of Technology, Atlanta, GA 30332 USA
        {\tt\footnotesize \{zhigen.zhao, mleggiero, ychen3302, hliu852, ywu3057, hxue37, sirui.zhan, yzhao301\}@gatech.edu}}}%
\thanks{\changed{Project website: \texttt{https://vqactflow.github.io/vqactflow/}}}
}
\begin{document}

\maketitle

\begin{abstract}
Multi-task robot manipulation policies are challenging to learn from demonstration because traditionally a single network must select among qualitatively different action modes from a multimodal demonstration distribution, conditioned on language and visual context. A wrong mode selection means executing the wrong task or an action infeasible in the scene. Tokenizing continuous actions into a learned discrete codebook separates these modes at the representation level, offering structural advantages for multi-task learning. We propose VQActFlow, a multi-task manipulation policy that tokenizes action chunks and generates code sequences via Variational Flow Matching. VQActFlow maintains an explicit preference over action modes throughout generation. Inference-time guidance acts on this preference to steer mode commitment. We instantiate this with classifier-free guidance over language conditioning, which steers the policy toward the instructed action mode, and a learned codebook critic that supplies a complementary feasibility signal. We evaluate VQActFlow on three platforms: the LIBERO simulation benchmarks, a Unitree G1 humanoid performing whole-body pick-and-place, and an ALOHA-style bimanual platform performing contact-rich tasks. Across these benchmarks, VQActFlow outperforms both continuous and discrete baselines.

\end{abstract}


\section{INTRODUCTION}
\IEEEPARstart{R}{obot} policies are increasingly expected to handle multiple manipulation tasks within a single model~\cite{bharadhwaj2024roboagent, ma2024hierarchical, lee2024behavior, zhao2024survey}, taking a language instruction and visual observation as input and dispatching to one of many learned behaviors at deployment time. Demonstration data for this setting is multimodal: the same observation recurs under different instructions and is followed by qualitatively different actions. A policy trained on such data must select the instructed action mode from this multimodal distribution, distinguishing modes that differ in target object, contact strategy, or bimanual coordination. When the observation is ambiguous, a wrong action mode selection is not a small numerical error but the robot carrying out a coherent behavior for the wrong task or performing the right task in an infeasible manner. The policy must therefore commit to the action mode consistent with the instruction and feasible in the scene.

Diffusion and flow matching policies~\cite{chi2025diffusion, lipman2024flow, black2024pi_0} have become the prevailing paradigm for visuomotor manipulation, and classifier-free guidance (CFG)~\cite{ho2022classifier} and related steering methods~\cite{dhariwal2021diffusion, janner2022planning} are the standard tools for biasing them at inference. These policies generate in a continuous action or latent space where action modes are not explicitly represented, so guidance reshapes a single continuous distribution rather than selecting among discrete action modes. Vector-quantized codebooks~\cite{van2017neural, lee2024behavior, wu2025discrete} instead learn a finite vocabulary of action primitives whose discrete bottleneck separates modes explicitly, exposing the structure the multi-task setting requires. Discrete Policy~\cite{wu2025discrete}, the closest prior work to ours, tokenizes actions with a VQ-VAE but generates codes by diffusion in a continuous space, recovering discrete indices by nearest-neighbor lookup only at the end of sampling, abandoning the discrete structure during generation. The policy is left with no explicit representation of which mode it is committing to, which is exactly what the multi-task setting needs to resolve ambiguous observations toward the instructed task.

We propose VQActFlow, a multi-task manipulation policy that makes action mode commitment \textit{explicit}: it maintains a preference over the discrete action modes throughout generation, so the mode commitment is represented at every step rather than recovered only at the end.
VQActFlow generates action tokens via Variational Flow Matching (VFM)~\cite{eijkelboom2024variational, maticsan2025purrception}, transporting Gaussian noise toward the embeddings of a frozen Vector-Quantized Variational Autoencoder (VQ-VAE) codebook under categorical supervision over code indices, so that it produces logits over codes at every integration step as an explicit preference over action modes. Inference-time guidance acts on this preference by redirecting probability toward the correct action mode, so the policy's commitment can be steered at deployment rather than fixed at training time.
We instantiate this guidance interface with two mechanisms. CFG amplifies the difference between task-conditioned and unconditional predictions to steer the policy toward the language-instructed action mode, and a learned codebook critic, trained via contrastive learning against structurally diverse negative examples, pushes the policy toward action modes feasible in the current scene. Both mechanisms act through the same interface and contribute complementary signals at each integration step.

Our contributions are as follows:
\begin{enumerate}
\item We propose VQActFlow, a multi-task manipulation policy that generates action tokens via Variational Flow Matching, maintaining an explicit preference over action modes throughout generation.
\item We show that guidance through this interface effectively steers action-mode selection at inference time, and introduce a novel contrastively-trained codebook critic that supplies a feasibility signal, operating on its own or alongside CFG.
\item We evaluate on the LIBERO benchmarks and on bimanual and humanoid hardware tasks, demonstrating improved performance over both continuous and discrete baselines.
\end{enumerate}

\section{Related Work}
\textbf{Diffusion and flow-based robot policies.}
Diffusion models~\cite{ho2020denoising} generate samples by iteratively denoising Gaussian noise toward the data distribution. Flow matching~\cite{lipman2024flow} reframes this as transporting noise along ODE trajectories defined by learned velocity fields. Diffusion Policy~\cite{chi2025diffusion} applies the diffusion paradigm to robot action chunks, modeling multi-step action distributions via DDPM denoising and achieving strong visuomotor control on contact-rich manipulation tasks. DiffuseLoco~\cite{huang2025diffuseloco} extends this paradigm to legged locomotion with receding-horizon control from offline multi-skill datasets, while Hybrid Diffusion~\cite{hoeg2026hybrid} jointly produces symbolic and continuous plans via coupled discrete-continuous denoising. $\pi_0$~\cite{black2024pi_0} builds a flow-matching policy with a VLM backbone for generalist manipulation. These methods operate in continuous action spaces, where guidance operates on continuous actions rather than discrete action modes. Discrete formulations of flow matching~\cite{gat2024discrete, campbell2024generative} transport probability over categorical states directly, whereas Variational Flow Matching~\cite{eijkelboom2024variational, maticsan2025purrception} retains a continuous transport, casting it as variational inference over categorical posteriors. To our knowledge, VFM has not been applied to robot policy learning.

\textbf{Robot action tokenization.}
OpenVLA~\cite{kim2024openvla} discretizes actions via per-dimension scalar binning to reuse language model vocabularies, a simple scheme that sacrifices inter-dimension correlations. Discrete Policy~\cite{wu2025discrete} learns a VQ-VAE codebook over action sequences and shows that the learned codes cluster by task similarity, suggesting that VQ-VAE tokenization captures task-discriminative structure in multi-task action distributions. VQ-BeT~\cite{lee2024behavior} uses a residual VQ-VAE with hierarchical code prediction via classification heads and a continuous offset for refinement. FAST~\cite{pertsch2025fast} applies DCT-based frequency compression followed by byte-pair encoding, while OAT~\cite{liu2026oat} learns ordered tokens via finite scalar quantization and nested dropout.
\revised{Masked generative policies~\cite{zhuang2026masked, liang2026discrete} predict action tokens iteratively and re-mask uncertain ones, whereas VQActFlow transports a continuous state and guides its categorical endpoint posterior at each step.}

\revised{\noindent\textbf{Multi-modal imitation learning and action abstraction.}
Discrete formulations make demonstration modes explicit by classifying over action bins~\cite{shafiullah2022behavior, lee2024behavior} or skill codes~\cite{garg2022lisa, mete2024quest}, but commit to a mode in a single prediction. CFG applied to a continuous action denoiser has been observed to bias action selection and reduce task completion~\cite{pearce2023imitating}. VQActFlow instead carries the categorical inside the flow, where guidance steers it at every integration step.}

\textbf{Guidance for generative policies.}
Classifier guidance~\cite{dhariwal2021diffusion} steers diffusion sampling with a trained classifier's gradients; CFG~\cite{ho2022classifier} avoids a separate classifier by interpolating conditional and unconditional scores. Guided Flows~\cite{zheng2023guided} extends guidance to flow matching with faster sampling than diffusion. In robotics, Diffuser~\cite{janner2022planning} applies reward guidance for trajectory planning, SafeDiffuser~\cite{xiao2025safediffuser} enforces safety constraints via control barrier functions, physics-informed diffusion~\cite{bastek2025physics} incorporates first-principles losses, and model-based diffusion~\cite{pan2024model} leverages learned dynamics models for trajectory optimization.
\revised{Beyond robotics, contrastive energy prediction~\cite{lu2023contrastive} learns a guidance energy contrastively, as our critic does, but steers a continuous score, and guidance for discrete-state diffusion and flows~\cite{nisonoff2025unlocking} corrects jump-process transition rates, which the continuous transport of VQActFlow does not require.}

\section{Preliminaries}
\label{sec:preliminaries}

\subsection{Vector-Quantized Variational Autoencoder}
\label{sec:prelim_vqvae}

A VQ-VAE~\cite{van2017neural} compresses a continuous input into a sequence of indices into a learned codebook $\mathcal{C} = \{e_k\}_{k=1}^K$, where each $e_k \in \mathbb{R}^{d_e}$ is a codebook embedding. An encoder $\mathcal{E}$ produces a sequence of continuous latents $z \in \mathbb{R}^{L \times d_e}$, each of which is quantized to its nearest codebook entry:
\begin{equation}
k_l^* = \arg\min_{k} \|z^{(l)} - e_k\|_2, \quad z_q^{(l)} = e_{k_l^*}.
\label{eq:vq_quantize}
\end{equation}
A decoder $\mathcal{D}$ reconstructs the input from the quantized embeddings $z_q$. The codebook is trained jointly with the encoder and decoder by combining a reconstruction loss with a commitment loss that pulls the encoder output toward its assigned codebook entry, with the codebook itself typically updated by exponential moving average~\cite{van2017neural, esser2021taming}. The result is a discrete bottleneck: any input is represented exactly by its sequence of $L$ codebook indices.

\subsection{Flow Matching}
\label{sec:prelim_fm}

Flow matching~\cite{lipman2024flow} learns a generative model by training a neural network to predict a velocity field that transports samples from a simple source distribution $p_0$ (typically $\mathcal{N}(0, I)$) to a data distribution $p_1$. Given a data sample $x_1 \sim p_1$, a conditional probability path $p_t(x \mid x_1)$ connects $p_0$ to $x_1$, and is characterized by a conditional velocity field $u_t(x \mid x_1)$. A velocity network $v_\theta(x_t, t)$ is trained to regress this conditional velocity:
\begin{equation}
\mathcal{L}_{\text{CFM}} = \mathbb{E}_{t,\, x_1,\, x_t \sim p_t(\cdot\,|\,x_1)}\!\left[\|v_\theta(x_t, t) - u_t(x_t \mid x_1)\|_2^2\right].
\label{eq:cfm_loss}
\end{equation}
A common choice is the optimal-transport conditional path, defined by sampling $x_0 \sim p_0$ and setting
\begin{equation}
x_t = (1-t)\,x_0 + t\,x_1, \quad u_t(x_t \mid x_1) = x_1 - x_0,
\label{eq:ot_path}
\end{equation}
which gives straight-line interpolants and a constant conditional velocity. At inference, the learned velocity field is integrated from $t{=}0$ to $t{=}1$, typically by Euler steps, to transport a fresh source sample $x_0 \sim p_0$ to a sample from the learned data distribution. Conditioning on context $c$ is incorporated by passing $c$ into $v_\theta(x_t, t, c)$ during both training and sampling.

\begin{figure*}[t]
\centering
\includegraphics[width=0.95\textwidth]{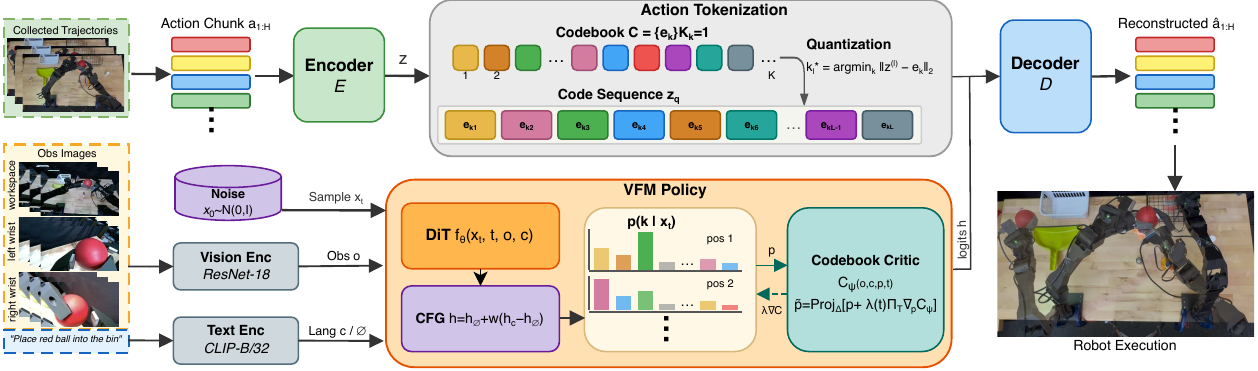}
\caption{VQActFlow framework. \textbf{Stage~1:} a VQ-VAE encoder tokenizes action chunks into discrete codebook indices, and a decoder reconstructs actions from the quantized embeddings. \textbf{Stage~2:} a VFM policy transports Gaussian noise toward codebook embeddings, maintaining an explicit preference over action modes that inference-time guidance steers, CFG toward the instructed task and a codebook critic toward feasible modes.}
\label{fig:framework}
\end{figure*}

\subsection{Classifier-Free Guidance}
\label{sec:prelim_cfg}

CFG~\cite{ho2022classifier} steers a conditional generative model toward objectives more strongly aligned with the conditioning $c$ without requiring an auxiliary classifier. During training, the conditioning is replaced by a learned null embedding $\varnothing$ with probability $p_{\text{drop}}$, so the same network $v_\theta$ learns to predict both the conditional output $v_\theta(x_t, t, c)$ and the unconditional output $v_\theta(x_t, t, \varnothing)$. At inference, the two outputs are linearly extrapolated to produce a guided output
\begin{equation}
\tilde{v} = v_\varnothing + w \cdot (v_c - v_\varnothing),
\label{eq:cfg_general}
\end{equation}
where $w \geq 1$ is the guidance weight. Setting $w{=}1$ recovers conditional sampling; $w > 1$ pushes the output further along the differential between conditional and unconditional predictions, sharpening the influence of $c$ on the generated sample.

\section{Methods}

\revised{Fig.~\ref{fig:framework} summarizes the framework: a VQ-VAE tokenizes action chunks (Sec.~\ref{sec:vqvae}), a VFM policy generates the codes (Sec.~\ref{sec:vfm_policy}), and CFG (Sec.~\ref{sec:guidance}) and a codebook critic (Sec.~\ref{sec:critic_guidance}) guide the generation.}

\subsection{VQ-VAE Action Tokenization}
\label{sec:vqvae}

We adopt the VQ-VAE framework from Sec.~\ref{sec:prelim_vqvae} to tokenize action chunks. Given an action chunk $a_{1:H} \in \mathbb{R}^{H \times d_a}$ spanning $H$ timesteps with $d_a$ action dimensions, a temporal encoder $\mathcal{E}$ maps the chunk to a sequence of continuous latents $z = \mathcal{E}(a_{1:H}) \in \mathbb{R}^{L \times d_e}$, where $L = H / 2^S$ is the compressed sequence length determined by $S$ temporal downsampling stages and $d_e$ is the codebook embedding dimension. Each latent is quantized to its nearest codebook entry as in Eq.~\ref{eq:vq_quantize}, and a temporal CNN decoder $\mathcal{D}$ reconstructs the action chunk from the quantized embeddings, so each output timestep depends on a local neighborhood of codes rather than on a single code in isolation. Each action chunk is thus represented by $L$ indices into a codebook of size $K$, with $L \ll H$. After training, the VQ-VAE is frozen and used to produce supervision targets for the VFM policy in Sec.~\ref{sec:vfm_policy}.

\subsection{Variational Flow Matching Policy}\label{sec:vfm_policy}
To preserve the discrete structure of the codebook during generation, VFM trains the network on a cross-entropy \revised{(CE)} over codebook indices rather than an L2 regression toward codebook embeddings, treating the $K$ codes as competing action modes and producing an explicit preference over them at every integration step. \revised{Unlike Discrete Policy~\cite{wu2025discrete}, which recovers the code assignment only at the final step, this preference is available throughout generation, and unlike a fully discrete process~\cite{campbell2024generative}, the transport keeps a continuous state for guidance to adjust (Table~\ref{tab:method_comparison}).}

\begin{table}[t]
\centering
\caption{\revised{Comparison of VQActFlow with related policies.}}
\label{tab:method_comparison}
\footnotesize
\setlength{\tabcolsep}{2.5pt}
\revised{
\begin{tabular}{@{}lcccc@{}}
\toprule
Method & \begin{tabular}[b]{@{}c@{}}Action\\space\end{tabular} & \begin{tabular}[b]{@{}c@{}}Generative\\process\end{tabular} & \begin{tabular}[b]{@{}c@{}}Training\\loss\end{tabular} & \begin{tabular}[b]{@{}c@{}}Guidance\\acts on\end{tabular} \\
\midrule
CFM & Continuous & Flow & L2 on velocity & Velocity \\
VQ-BeT & Residual VQ & Autoregressive & Focal + L1 & -- \\
Discrete Policy & VQ-VAE & Diffusion & L2 on noise & Embedding \\
VQActFlow & VQ-VAE & Flow & CE on codes & Posterior \\
\bottomrule
\end{tabular}
}
\end{table}

\revised{
\textbf{Training.} Given a demonstration sample, the frozen VQ-VAE encoder (Sec.~\ref{sec:vqvae}) produces target embeddings $x_1 = z_q \in \mathbb{R}^{L \times d_e}$ and ground-truth indices $\{k_l^*\}_{l=1}^L$. We sample source noise $x_0 \sim \mathcal{N}(0, I) \in \mathbb{R}^{L \times d_e}$ and a timestep $t \sim \mathcal{U}(0, 1)$, and construct the optimal-transport interpolant $x_t = (1-t)\,x_0 + t\,x_1$ from Eq.~\ref{eq:ot_path}. For this path the conditional velocity is $u_t(x_t \mid x_1) = (x_1 - x_t)/(1-t)$, and the marginal velocity that transports $p_0$ to $p_1$ is its posterior expectation,
\begin{equation}
u_t(x) = \frac{\mathbb{E}\left[x_1 \mid x_t = x\right] - x}{1 - t},
\label{eq:marginal_velocity}
\end{equation}
which depends on the posterior $p(x_1 \mid x_t)$ through its mean alone. Variational flow matching~\cite{eijkelboom2024variational} fits this posterior with a model $q_\theta(x_1 \mid x_t)$ by maximizing the expected log-likelihood $\mathbb{E}[\log q_\theta(x_1 \mid x_t)]$. Because each $x_1^{(l)}$ takes values on the codebook $\{e_1, \ldots, e_K\}$, $q_\theta$ is a product of categorical distributions over code indices, and the factorization does not affect the mean in Eq.~\ref{eq:marginal_velocity}. The policy network $f_\theta$ therefore processes $x_t$ conditioned on $t$, an observation encoding $o$, and language conditioning $c$, producing logits over the codebook at every position:
\begin{equation}
h^{(l)} = f_\theta(x_t, t, o, c)^{(l)} \in \mathbb{R}^{K}, \quad l = 1, \ldots, L.
\label{eq:vfm_logits}
\end{equation}
With $q_\theta(k \mid x_t)^{(l)} = \exp(h_k^{(l)}) / \sum_j \exp(h_j^{(l)})$, the negative log-likelihood of the ground-truth indices is the cross-entropy
\begin{equation}
\mathcal{L}_{\text{VFM}} = -\sum_{l=1}^{L} \left( h^{(l)}_{k_l^*} - \log \sum_{j=1}^{K} \exp h^{(l)}_j \right).
\label{eq:vfm_loss}
\end{equation}
The L2 loss in Eq.~\ref{eq:cfm_loss} is minimized by the same marginal velocity, Eq.~\ref{eq:marginal_velocity}; the two objectives differ in whether $\mathbb{E}[x_1 \mid x_t]$ is parameterized directly or through a distribution over modes.
}

\revised{
\textbf{Inference.} At each integration step, the velocity is obtained by substituting the model's posterior mean for $\mathbb{E}[x_1 \mid x_t]$ in the marginal velocity above. Let $p_k^{(l)} = \exp(h_k^{(l)}) / \sum_j \exp(h_j^{(l)})$. Since $x_1^{(l)}$ takes values on the codebook, the posterior mean under $q_\theta$ is
\begin{equation}
\mu_1^{(l)} = \mathbb{E}_{q_\theta}\!\left[x_1^{(l)} \mid x_t\right] = \sum_{k=1}^{K} p_k^{(l)} \, e_k,
\label{eq:posterior_mean}
\end{equation}
which gives the velocity
\begin{equation}
v_\theta(x_t, t) = \frac{\mu_1 - x_t}{1 - t}.
\label{eq:velocity}
\end{equation}
}

The velocity points toward the codebook-weighted average of the action modes currently favored by the policy, with $p_k^{(l)}$ controlling the relative weight on each candidate. As the policy concentrates probability on a single mode during integration, this average becomes dominated by the embedding of the favored mode.
Starting from $x_0 \sim \mathcal{N}(0, I)$, we integrate $\dot{x} = v_\theta(x, t)$ from $t{=}0$ to $t{=}1$ using Euler steps. The terminal state is quantized position-wise, $k_l^* = \arg\min_k \|x_1^{(l)} - e_k\|_2$, and the resulting code sequence is decoded to actions $a_{1:H} = \mathcal{D}(e_{k_1^*}, \ldots, e_{k_L^*})$.

\textbf{Architecture.} The policy network $f_\theta$ is a Diffusion Transformer (DiT)~\cite{peebles2023scalable} backbone with cross-attention for visual and text conditioning and adaptive layer normalization for the flow-matching timestep embedding. The visual encoder is a ResNet-18~\cite{he2016deep} and the text encoder is a frozen CLIP-B/32~\cite{radford2021learning}.

\subsection{Classifier-Free Guidance for Language Conditioning}
\label{sec:guidance}

We apply CFG (Sec.~\ref{sec:prelim_cfg}) to the language conditioning $c$, with the network producing conditional logits $h_c = f_\theta(x_t, t, o, c)$ and unconditional logits $h_\varnothing = f_\theta(x_t, t, o, \varnothing)$. At each integration step, the two are linearly extrapolated on the logits rather than on the velocity:
\begin{equation}
\tilde{h} = h_\varnothing + w \cdot (h_c - h_\varnothing),
\label{eq:cfg_logits}
\end{equation}
and the guided probabilities are computed as
\begin{equation}
\tilde{p}_k^{(l)} = \frac{\exp(\tilde{h}_k^{(l)})}{\sum_{j=1}^{K} \exp(\tilde{h}_j^{(l)})},
\label{eq:cfg_probs}
\end{equation}
which replace $p$ in Eq.~\ref{eq:posterior_mean} to compute the posterior mean and velocity. The guided distribution biases the policy's mode preference toward codes consistent with the language conditioning, effectively sharpening which action mode the policy commits to under ambiguous observations.

\subsection{Codebook Critic for Scene-Conditioned Feasibility}
\label{sec:critic_guidance}

The codebook critic supplies a learned feasibility signal that biases the policy's mode preference toward actions feasible in the current scene. It operates on the categorical distribution over codes rather than on decoded actions, so each guidance step requires only one forward and backward pass through a small network, with no codebook decoding in the inner loop. Because the critic is structured independently of the policy, its objective can be adapted to different feasibility criteria without retraining the policy itself, and it is trained on demonstration data alone, with no simulator rollouts or policy samples required.

\textbf{Guided velocity.} During ODE integration, the critic gradient adjusts the policy's preference over action modes:
\begin{equation}
\tilde{p} = \text{Proj}_{\Delta}\!\left[p + \lambda(t) \cdot \Pi_T\!\left[\nabla_{p} C_\psi(o, c, p, t)\right]\right],
\label{eq:guided_velocity}
\end{equation}
where $p \in \mathbb{R}^{L \times K}$ is the current categorical distribution over codebook entries, $\Pi_T[g] = g - \bar{g}$ removes the component-wise mean to keep the gradient tangent to the simplex, and $\text{Proj}_{\Delta}$ projects onto the probability simplex. The guidance weight adapts to the entropy of the current distribution:
\begin{equation}
\lambda(t) = \lambda_{\max}\!\left(1 - \frac{H(p)}{\log K}\right),
\label{eq:entropy_weight}
\end{equation}
where $H(p) = -\sum_k p_k \log p_k$. \revised{Early posteriors are near-uniform and reflect the noise rather than the data, so $\lambda(t)$ is small while $p$ is diffuse and grows as $p$ concentrates.}

\textbf{Critic architecture and training.}
The critic $C_\psi(o, c, p, t)$ is a transformer encoder that maps a visual observation $o$, language conditioning $c$, sequence of $L$ simplex positions $p \in \mathbb{R}^{L \times K}$, and flow timestep $t$ to a scalar feasibility score. The visual and language inputs are fused by an MLP into a conditioning token that is prepended to the code sequence. Each $p^{(l)}$ is linearly projected to $d_{\text{model}}$ and combined with a learned positional embedding and a sinusoidal encoding of $t$. A pre-norm transformer encoder processes the sequence, and a mean-pooled representation is projected to a scalar by an MLP head.

Training uses an InfoNCE loss~\cite{oord2018representation} that contrasts a positive $(o, c, p)$ triple against negatives constructed to violate distinct components of trajectory validity:
\begin{itemize}
\item \textit{Temporal shuffle}: permutes the order of the code sequence.
\item \textit{Random code replacement}: replaces a fraction of positions with uniformly sampled codebook entries.
\item \textit{Wrong observation}: pairs a valid code sequence with a mismatched observation from another sample in the batch.
\item \textit{Cross-demonstration swap}: replaces the code sequence with one drawn from a different demonstration.
\item \textit{In-episode temporal shift}: replaces the code sequence with one from the same episode at a different timestep.
\item \textit{Wrong language, correct vision}: permutes the language conditioning across the batch while keeping vision and codes paired.
\item \textit{Wrong vision, correct language}: permutes the visual conditioning while keeping language and codes paired.
\end{itemize}
\revised{Together these negatives, grouped as conditioning (wrong observation, language, vision), code-replacement (random replacement, cross-demonstration swap), and temporal (shuffle, shift) negatives, expose the critic to violations of observation and language grounding, action correctness, and temporal coherence.}

All samples are noised to random flow timesteps $t \sim \mathcal{U}(t_{\min}, t_{\max})$ via the interpolation $x_t = (1-t)/K + t \cdot x_{\text{clean}}$, exposing the critic to a range of distributional sharpness.

\section{Experiments}
\label{sec:experiments}

We evaluate VQActFlow policy in simulation on the LIBERO benchmark suites~\cite{liu2023libero} and on hardware with bimanual manipulation tasks on a set of ALOHA-style arms and a Unitree G1 humanoid robot. We evaluate VQActFlow against three claims. First, the categorical interface enables effective inference-time guidance for action-mode selection (Sec.~\ref{sec:libero_goal}, Sec.~\ref{sec:libero_90}, Sec.~\ref{sec:g1_results}). Second, the contrastively-trained codebook critic supplies a useful feasibility signal, operating independently or alongside CFG (Sec.~\ref{sec:libero_90}, Sec.~\ref{sec:bimanual}). Third, VQActFlow achieves the highest success rates among comparable continuous and discrete baselines across LIBERO-90, bimanual hardware, and humanoid hardware (Sec.~\ref{sec:libero_90}, Sec.~\ref{sec:bimanual}, Sec.~\ref{sec:g1_results}).

\begin{figure}[t]
\centering
\includegraphics[width=0.8\columnwidth]{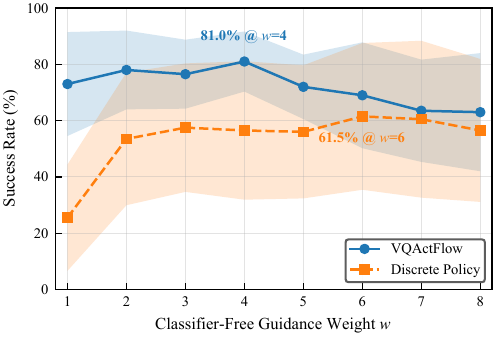}
\caption{LIBERO-Goal success rate vs.\ CFG weight. VQActFlow peaks at $w{=}4$ and then degrades at higher weights. Discrete Policy is shown for reference under matched conditions. Shaded: $\pm 1$ std across 10 tasks.}
\label{fig:cfg_sweep}
\end{figure}

Throughout the experiments, we use a flat VQ-VAE codebook of size $K{=}512$ with $S{=}2$ temporal downsampling stages. The VFM policy backbone is a DiT with $12$ layers, hidden size $d_{\text{model}}{=}1024$, and $8$ attention heads. The codebook critic is a smaller transformer with $3$ layers, hidden size $256$, and $4$ attention heads. Both the policy and critic use a frozen CLIP-B/32~\cite{radford2021learning} text encoder; the policy and critic visual encoder is shared and held frozen during critic training.
The implementation uses the \texttt{flow\_matching}~\cite{lipman2024flow} and \texttt{LeRobot}~\cite{cadene2024lerobot} libraries.
All baseline policies are trained and evaluated under matched protocol: identical vision and text encoders, training dataset, number of training steps, and initialization seeds for simulation benchmarks. \revised{The Discrete Policy~\cite{wu2025discrete} baseline shares the VQ-VAE weights and DiT backbone of VQActFlow.} \revised{Guidance weights are chosen by a coarse sweep of success rate per benchmark, with no per-task tuning, and $\lambda_{\max}{=}1$ (denoted $\lambda$) is used wherever the critic is enabled.}

\subsection{CFG Evaluation on LIBERO-Goal}
\label{sec:libero_goal}
\revised{LIBERO-Goal evaluates how CFG steers action-mode selection when all tasks share one scene and only the instruction identifies the task.}
We sweep the CFG weight $w \in \{1, 2, \ldots, 8\}$ on LIBERO-Goal with 20 rollouts per task and compare against Discrete Policy under matched conditions. VQActFlow improves from $73.0\%$ unguided to $81.0\%$ at $w{=}4$, then degrades monotonically to $63.0\%$ at $w{=}8$ as per-task variance widens, as seen in Fig.~\ref{fig:cfg_sweep}. Discrete Policy peaks at $w{=}6$ with $61.5\%$, requiring stronger guidance to reach a substantially lower ceiling. VQActFlow's tighter standard deviation band further indicates that the gain is consistent across tasks.

\begin{figure}[t]
\centering
\setlength{\tabcolsep}{0pt}
\begin{tabular}{c@{\hspace{4pt}}c}
\includegraphics[height=0.4404\columnwidth,trim=8 0.4 6 1.3,clip]{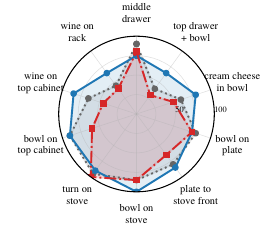} &
\includegraphics[height=0.4404\columnwidth,trim=1 0.5 1 0.5,clip]{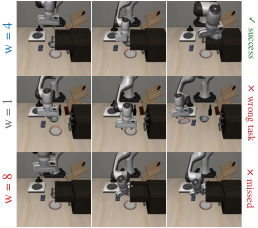} \\[-1pt]
{\footnotesize (a)} & {\footnotesize (b)}
\end{tabular}
\caption{\revised{CFG weight effects on LIBERO-Goal. (a) Per-task success at $w{=}1$ (unguided, grey), $w{=}4$ (peak, blue), and $w{=}8$ (over-guided, red). (b) Rollouts of ``open the top drawer and put the bowl inside'': success ($w{=}4$), wrong-task failure ($w{=}1$), missed-grasp failure ($w{=}8$).}}
\label{fig:libero_goal_cfg}

\end{figure}

The aggregate response masks substantial per-task heterogeneity (\revised{Fig.~\ref{fig:libero_goal_cfg}(a)}). Tasks with significant unguided headroom gain the most from CFG: ``open the top drawer and put the bowl inside'' ($40 \to 65\%$ at $w{=}4$), ``put the wine bottle on top of the cabinet'' ($65 \to 85\%$), and ``put the cream cheese in the bowl'' ($60 \to 80\%$). Near-ceiling tasks regress: ``open the middle drawer'' drops from $90\%$ to $75\%$ at $w{=}4$. Rollout inspection in \revised{Fig.~\ref{fig:libero_goal_cfg}(b)} reveals two distinct failure regimes. At low $w$, the policy executes coherent actions on the wrong object, indicating that the unconditional marginal $h_\varnothing$ is insufficiently suppressed. At high $w$, the policy reaches the correct object but misses the grasp, suggesting that aggressive amplification of $(h_c - h_\varnothing)$ degrades execution precision.
Optimal $w$ thus trades off suppression of wrong-task modes against degradation of execution.

\subsection{Multi-task Scaling on LIBERO-90}
\label{sec:libero_90}
LIBERO-90 evaluates VQActFlow on 90 multi-task scenarios that test both the guidance interface at scale and the contribution of the codebook critic. All baselines are trained from scratch under matched conditions to isolate the action-generation mechanism, so vision-language-action (VLA) models~\cite{kim2024openvla, black2024pi_0} with large pretraining datasets fall outside this comparison. We report \revised{the mean and standard deviation of success rate over 5 independent evaluation seeds with 20 rollouts per task, and differences with 95\% Welch $t$-intervals over seed means,} for VQActFlow with CFG alone, critic guidance alone, and both combined, along with baselines, as seen in Table~\ref{tab:libero90}. VQActFlow with $w{=}2$ and $\lambda{=}1.0$ reaches \revised{$79.3\pm1.1\%$}, surpassing Continuous Flow Matching DiT (CFM) \revised{with no CFG}, MT-ACT~\cite{bharadhwaj2024roboagent}, and all VQ-based baselines.

\revised{CFG is effective for both continuous and categorical interfaces, improving CFM by $+2.1\%$ $[0.6, 3.7]$ and VQActFlow by $+4.8\%$ $[2.9, 6.6]$ at $w{=}2$. Critic guidance alone adds $+2.4\%$ $[0.4, 4.3]$, and on top of CFG a further $+2.7\%$ $[0.7, 4.6]$, so the two signals are largely non-redundant. CFG sharpens task identity through the contrast between $h_c$ and $h_\varnothing$, while the critic's gain comes from its code-replacement negatives (Table~\ref{tab:critic_ablation}).} Among baselines with VQ-based latent action spaces, VQ-BeT achieves \revised{$23.0\pm1.3\%$}, and Discrete Policy\revised{, the closest published discrete-token method,} with CFG trails VQActFlow at matched guidance weight by \revised{$15.8\%$ $[13.9, 17.7]$}.

\begin{table}[t]
\centering
\caption{\revised{LIBERO-90 success rate (\%) for different policies.}}
\label{tab:libero90}
\revised{%
\begin{tabular}{llccc}
\hline
\textbf{Method} & \textbf{Type} & \textbf{$w$} & \textbf{$\lambda$} & \textbf{Success} \\
\hline
MT-ACT~\cite{bharadhwaj2024roboagent} & Continuous & -- & -- & $72.2_{\pm0.2}$ \\
CFM & Continuous & -- & -- & $76.2_{\pm1.2}$ \\
CFM & Continuous & 2 & -- & $78.3_{\pm0.9}$ \\
VQ-BeT~\cite{lee2024behavior} & VQ-based & -- & -- & $23.0_{\pm1.3}$ \\
Discrete Policy~\cite{wu2025discrete} & VQ-based & 1 & -- & $49.8_{\pm1.2}$ \\
Discrete Policy & VQ-based & 2 & -- & $60.8_{\pm1.0}$ \\
\hline
VQActFlow & VQ-based & 1 & 0 & $71.8_{\pm0.7}$ \\
VQActFlow & VQ-based & 1 & 1 & $74.2_{\pm1.5}$ \\
VQActFlow & VQ-based & 2 & 0 & $76.6_{\pm1.5}$ \\
\textbf{VQActFlow} & \textbf{VQ-based} & \textbf{2} & \textbf{1} & $\mathbf{79.3_{\pm1.1}}$ \\
\hline
\end{tabular}%
}
\end{table}
Table~\ref{tab:libero90_codebook} ablates the codebook size $K$ on LIBERO-90 with guidance disabled. Success peaks at $K{=}512$: smaller codebooks limit vocabulary granularity, while larger ones enlarge the per-position classification space, making the categorical prediction harder for the policy to learn.

\begin{table}[t]
\centering
\caption{\revised{Ablation for codebook sizes with LIBERO-90 (\%).}}
\label{tab:libero90_codebook}
\revised{
\begin{tabular}{l c c c c}
\toprule
Codebook size & $128$ & $256$ & $512$ & $1024$ \\
\midrule
Success rate & $63.7_{\pm0.8}$ & $67.3_{\pm0.2}$ & $\mathbf{71.8_{\pm0.7}}$ & $70.6_{\pm0.7}$ \\
\bottomrule
\end{tabular}
}
\end{table}

\begin{figure}[t]
\centering
\includegraphics[width=0.8\columnwidth]{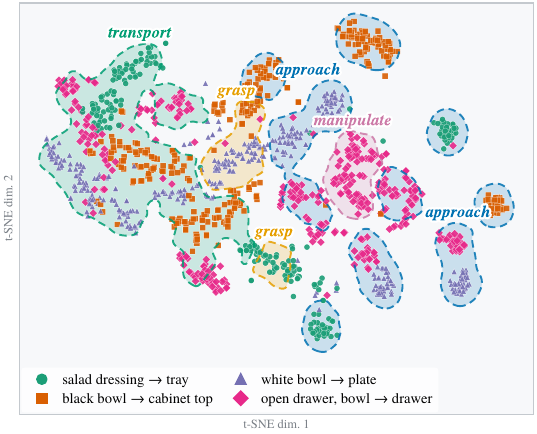}
\caption{\revised{The t-SNE visualization of action embeddings for 4 LIBERO-90 tasks, clustered by action semantics.}}
\label{fig:tsne_action_modes}
\end{figure}

\revised{
We embed the VQ-VAE latents of policy-predicted action chunks from successful LIBERO-90 rollouts with t-SNE (Fig.~\ref{fig:tsne_action_modes}), labeling each chunk by its replay-segmented subtask phase. Manipulation chunks are defined as articulated object operation, such as opening a drawer or turning a stove knob. Chunks cluster by action semantics, with approach, transport, and manipulation chunks each occupying their own regions regardless of task identity. This supports treating code selection as action-mode selection, the premise of both guidance mechanisms.
}

\revised{
We collect $4{,}500$ LIBERO-90 rollouts with the critic disabled and score every action chunk by the critic. We measure the ranking accuracy as the area under the receiver operating characteristic curve (AUROC), the probability that a successful chunk receives a higher score than a failed one. The critic reaches a ranking accuracy of $0.82$. Table~\ref{tab:critic_ablation} reports critics retrained with subsets of the negative families, evaluated by ranking accuracy, task discrimination (the AUROC of chunks scored under the correct vs.\ a wrong instruction), and LIBERO-90 success rate in a separate $4{,}500$-episode evaluation at $w{=}2$, $\lambda{=}1$. The conditioning negatives train task and scene consistency, and removing them lowers task discrimination from $0.97$ to $0.68$ but leaves the success rate unchanged. The code-replacement negatives contrast the demonstrated codes with codes from outside the episode at a fixed observation and instruction, and removing them leaves a critic that ranks rollouts as well as the full critic but lowers success rate to $66.4\%$.\looseness=-1
}

\begin{table}[t]
\centering
\caption{\revised{Critic negative-family ablation on LIBERO-90.}}
\label{tab:critic_ablation}
\revised{
\begin{tabular}{lccc}
\toprule
Critic negatives & Ranking & Task disc. & Success \\
\midrule
No critic           & ---  & ---  & 76.9 \\
All seven           & 0.82 & 0.97 & 78.8 \\
No conditioning     & 0.80 & 0.68 & 78.7 \\
No code replacement & 0.83 & 0.96 & 66.4 \\
\bottomrule
\end{tabular}
}
\end{table}

\subsection{Humanoid Whole-Body Manipulation}
\label{sec:g1_results}
\begin{figure}[t]
\centering
\setlength{\tabcolsep}{0pt}
\begin{tabular}{c@{\hspace{4pt}}c}
\includegraphics[height=0.4866\columnwidth]{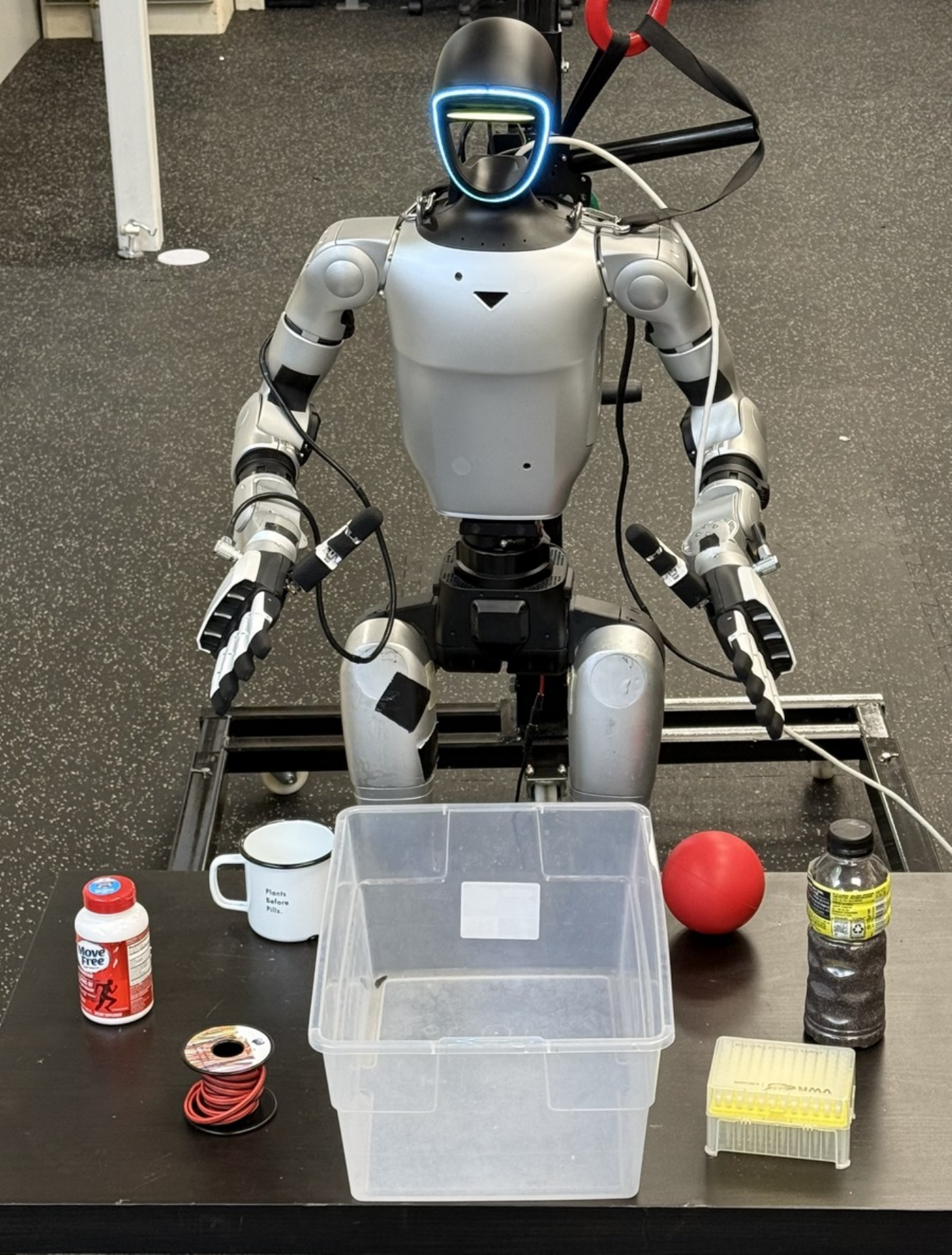} &
\includegraphics[height=0.4866\columnwidth]{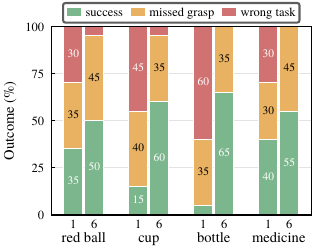} \\[-1pt]
{\footnotesize (a)} & {\footnotesize (b)}
\end{tabular}
\caption{\revised{G1 humanoid experiments. (a)~Multitask tabletop scene; the four tasks place the red ball, cup, bottle, or medicine bottle into the box. (b)~Rollout outcomes per task (20 trials each) at $w{=}1$ and $w{=}6$.}}
\label{fig:g1}
\end{figure}

We evaluate VQActFlow on a Unitree G1 humanoid equipped with two Inspire RH56DFTP dexterous hands, operated under whole-body control, shown in \revised{Fig.~\ref{fig:g1}(a)}. The proprioceptive state is $46$-dimensional and the action space is $47$-dimensional, with egocentric vision from an Intel RealSense D435i head camera at $424 \times 240$. Four pick-and-place tasks place one of a red ball, a cup, a bottle, or a medicine bottle into a shared bin \revised{(Fig.~\ref{fig:g1}(a))}. The four tasks share a workspace and differ only in target object and language instruction, isolating language-conditioned task disambiguation as the sole source of cross-task variation. We collect $140$ VR-teleoperated demonstrations per task using TWIST2~\cite{ze2025twist2} and roll out $20$ trials per task at $w{=}1$ and $w{=}6$, scoring each as success, missed grasp (correct target, failed execution), or wrong task (incorrect target).

CFG raises aggregate success from $23.8\%$ to $57.5\%$. A more detailed breakdown in \revised{Fig.~\ref{fig:g1}(b)} shows all four tasks improving with CFG: red ball ($+15\%$), cup ($+45\%$), bottle ($+60\%$), and medicine ($+15\%$).
The gain is driven almost entirely by the correction of wrong-task errors, which fall from $41.3\%$ to $2.5\%$, while the missed-grasp rate is approximately unchanged. This asymmetric gain reflects CFG's training objective: conditioning dropout teaches the policy to encode task identity in $h_c - h_\varnothing$, so guidance improves task disambiguation but not execution quality. The result mirrors the failure-mode dichotomy in simulation.

\subsection{Bimanual Manipulation}
\label{sec:bimanual}

\begin{figure}[t]
  \centering
  \includegraphics[width=0.85\columnwidth]{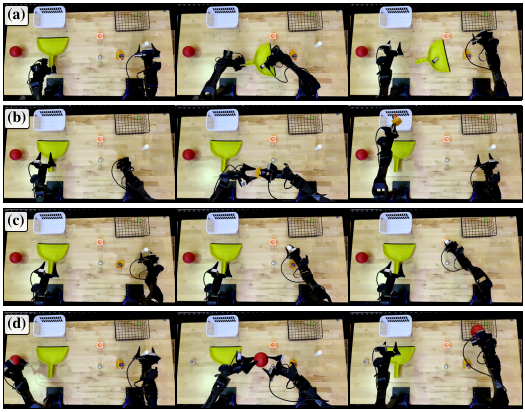}
\caption{Bimanual manipulation evaluation tasks.
  (a)~Sweep the battery into the dust pan.
  (b)~Place the rubber duck into the white bin.
  (c)~Place the cylinder into the hole.
  (d)~Place the red ball into the black bin.}
  \label{fig:bimanual_tasks}
\end{figure}

We evaluate VQActFlow on a bimanual platform of two ALOHA-style arms. Visual observations stream from three Intel RealSense D405 cameras: one wrist-mounted camera per arm and one fixed workspace camera. \revised{All task objects share one tabletop scene.} We design four contact-rich manipulation tasks (Fig.~\ref{fig:bimanual_tasks}): (1)~sweeping a battery into a dust pan, (2)~placing a rubber duck into a white bin, (3)~inserting a cylinder into a hole, and (4)~placing a red ball into a black bin. Tasks 1, 2, and 4 require coordinated use of both arms, while 3 is a single-arm peg-in-hole task that challenges insertion accuracy. We collect $100$ demonstration episodes per task via VR teleoperation using XRoboToolkit~\cite{zhao2026xrobotoolkit}.

VQActFlow is compared against Discrete Policy \revised{over 5 evaluation sessions of 20 trials per task}. The result is shown in Table~\ref{tab:bimanual_results}. A CFM baseline with the same DiT backbone is excluded from the table because it failed to complete any task; we examine this failure at the end of the section.

\revised{VQActFlow outperforms Discrete Policy with guidance. Without guidance the gap ($+6.2\%$, $[-1.0\%, 13.4\%]$) is within session variance; at $w{=}4$ VQActFlow reaches $78.5\%$ against $70.8\%$ ($+7.7\%$, $[2.0\%, 13.4\%]$), and the full configuration with $w{=}4$ and $\lambda{=}1$ reaches $82.5\%$ ($+11.7\%$, $[7.4\%, 16.0\%]$). Critic guidance improves VQActFlow on every task at both guidance weights, raising the average from $67.0\%$ to $72.8\%$ at $w{=}1$ and from $78.5\%$ to $82.5\%$ at $w{=}4$. Neither average gain is significant at this trial budget ($[-0.3\%, 11.9\%]$ and $[-1.7\%, 9.7\%]$), so we treat the hardware result as corroborating the critic gain measured on LIBERO-90.}

\begin{table}[t]
\centering
\caption{\revised{Success rates (\%) on bimanual manipulation tasks.}}
\label{tab:bimanual_results}
\footnotesize
\setlength{\tabcolsep}{2.5pt}
\renewcommand{\arraystretch}{1.15}
\revised{
\begin{tabular}{@{}lccccccc@{}}
\toprule
Method & $w$ & $\lambda$ & Battery & Duck & Cylinder & Ball & Avg. \\
\midrule
Discrete Policy & 1 & -- & $53_{\pm7.6}$ & $54_{\pm8.9}$ & $68_{\pm4.5}$ & $68_{\pm4.5}$ & $60.8_{\pm5.0}$ \\
Discrete Policy & 4 & -- & $64_{\pm5.5}$ & $71_{\pm5.5}$ & $78_{\pm4.5}$ & $70_{\pm10.0}$ & $70.8_{\pm2.9}$ \\
\midrule
VQActFlow & 1 & 0 & $60_{\pm7.9}$ & $76_{\pm5.5}$ & $77_{\pm7.6}$ & $55_{\pm11.2}$ & $67.0_{\pm4.9}$ \\
VQActFlow & 1 & 1 & $67_{\pm2.7}$ & $77_{\pm4.5}$ & $84_{\pm9.6}$ & $63_{\pm11.0}$ & $72.8_{\pm2.7}$ \\
VQActFlow & 4 & 0 & $74_{\pm4.2}$ & $80_{\pm3.5}$ & $85_{\pm7.9}$ & $75_{\pm15.4}$ & $78.5_{\pm4.5}$ \\
\textbf{VQActFlow} & \textbf{4} & \textbf{1} & $\mathbf{76_{\pm4.2}}$ & $\mathbf{85_{\pm5.0}}$ & $\mathbf{90_{\pm6.1}}$ & $\mathbf{79_{\pm6.5}}$ & $\mathbf{82.5_{\pm3.0}}$ \\
\bottomrule
\end{tabular}
}
\end{table}

\revised{On a workstation with an AMD Ryzen 7 5800X3D CPU and an NVIDIA RTX 5090 GPU, the average per-chunk inference latency over 50 trials is 155, 289, 217, and 350~ms for the unguided, CFG, critic, and combined configurations. CFG doubles the per-step DiT cost by evaluating the network twice, and the critic adds a smaller forward and backward pass. The full configuration completes well within the 1.28~s spanned by each action chunk, and asynchronous inference computes the next chunk during execution.}

\begin{figure}[t]
  \centering
  \includegraphics[width=0.8\columnwidth]{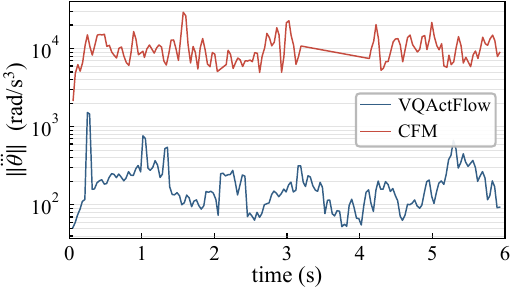}
\caption{Motion smoothness comparison between VQActFlow and CFM for bimanual manipulation hardware experiment.}
  \label{fig:bimanual_smoothness}
\end{figure}

A CFM policy with the same DiT backbone converges in training but fails all four tasks at evaluation. Rollout inspection reveals severe high-frequency oscillation in the commanded joint trajectories: Fig.~\ref{fig:bimanual_smoothness} shows peak joint jerk on the order of $10^4$\,rad/s$^3$, whereas VQActFlow remains near $2 \times 10^2$\,rad/s$^3$, roughly two orders of magnitude smoother. \revised{Representative rollouts are shown on the project website.}

\section{CONCLUSIONS}
\revised{VQActFlow keeps an explicit categorical preference over action codes throughout generation, a common interface for language and critic guidance. With CFG, it reaches higher success than Discrete Policy at matched training and guidance weight on LIBERO-Goal, LIBERO-90, and the bimanual tasks and peaks at a lower weight on LIBERO-Goal, though these comparisons do not isolate the interface alone.} The contrastively-trained codebook critic operates through the same interface and supplies a feasibility signal that CFG alone does not. VQActFlow deploys successfully on two distinct hardware platforms, a Unitree G1 humanoid and an ALOHA-style bimanual system, executing contact-rich tasks where continuous flow matching at matched capacity fails.

Several directions remain open. \changed{The effect of guidance is sensitive to the guidance weight, which we select per benchmark by a success-rate sweep, so deployment on a new robot platform may require further tuning through real-world experiments, which can be costly.} The VQ-VAE bottleneck imposes an information loss floor that ultimately bounds policy performance, and more expressive action tokenization schemes may relax this constraint while preserving the categorical interface. \changed{Another} direction is to use VFM as the action head of a VLA model, where the categorical output naturally interfaces with language-model token vocabularies.

\bibliographystyle{IEEEtran}
\bibliography{references.bib}

\end{document}